\documentclass[letterpaper, 10 pt, conference]{ieeeconf}  

\IEEEoverridecommandlockouts                              

\usepackage{graphicx} 
\usepackage{multirow}
\usepackage{amsmath} 
\usepackage{amssymb}  

\title{\LARGE \bf
LIWO: Lidar-Inertial-Wheel Odometry 
}

\author{Zikang~Yuan$^{1}$, Fengtian~Lang$^{2}$, Tianle~Xu$^{2}$ and Xin~Yang$^{2*}$
\thanks{This work is supported by the National Natural Science Foundation of China (62122029, 62061160490,  U20B2064).}
\thanks{$^{1}$Zikang~Yuan is with Institute of Artificial Intelligence, Huazhong University of Science and Technology, Wuhan, 430074, China. (E-mail: {\tt\small yzk2020@hust.edu.cn})}%
\thanks{$^{2}$Fengtian~Lang, Tianle~Xu and Xin~Yang (corresponding author) are with the Electronic Information and Communications, Huazhong University of Science and Technology, Wuhan, 430074, China. (E-mail: {\tt\small U201913666@hust.edu.cn; tianlexu@hust.edu.cn; xinyang2014@hust.edu.cn})}%
}

\begin{document}

\maketitle
\thispagestyle{empty}
\pagestyle{empty}

\begin{abstract}

LiDAR-inertial odometry (LIO), which fuses complementary information of a LiDAR and an Inertial Measurement Unit (IMU), is an attractive solution for state estimation. In LIO, both pose and velocity are regarded as state variables that need to be solved. However, the widely-used Iterative Closest Point (ICP) algorithm can only provide constraint for pose, while the velocity can only be constrained by IMU pre-integration. As a result, the velocity estimates inclined to be updated accordingly with the pose results. In this paper, we propose LIWO, an accurate and robust LiDAR-inertial-wheel (LIW) odometry, which fuses the measurements from LiDAR, IMU and wheel encoder in a bundle adjustment (BA) based optimization framework. The involvement of a wheel encoder could provide velocity measurement as an important observation, which assists LIO to provide a more accurate state prediction. In addition, constraining the velocity variable by the observation from wheel encoder in optimization can further improve the accuracy of state estimation. Experiment results on two public datasets demonstrate that our system outperforms all state-of-the-art LIO systems in terms of smaller absolute trajectory error (ATE), and embedding a wheel encoder can greatly improve the performance of LIO based on the BA framework.

\end{abstract}

\section{INTRODUCTION}
\label{Introduction}

As a widely-used solution for 6-degree of freedom (DOF) pose estimation and map reconstruction, LiDAR-inertial odometry (LIO) is a fundamental technique for many robotics applications, e.g., unmanned vehicle and automatic navigation. LIO combines the measurements from a three-dimension light detection and ranging (LiDAR) and an Inertial Measurement Unit (IMU) to estimate the state (i.e., pose and velocity) of the hardware platform in real time, and then utilizes the solved state to register the points of a new sweep into the map. According to the degree of coupling, existing LIO systems \cite{zhang2014loam, zhang2017low, shan2018lego, li2021towards, ye2019tightly, qin2020lins, shan2020lio, xu2021fast, xu2022fast, chen2022direct, yuan2022sr} can be divided into two groups: loose-coupled and tightly coupled.

The loose-coupled framework \cite{zhang2014loam, zhang2017low, shan2018lego} mainly uses IMU measurements to calibrate the motion distortion of LiDAR points, and provide motion priors for Iterative Closest Point (ICP) pose estimation. For instance, LOAM \cite{zhang2014loam} and LeGO-LOAM \cite{shan2018lego} have loosely-coupled IMU interfaces in their open source code, although they described their work as the LiDAR-only odometry and mapping system in the literature. However, calculating the motion priors of current time relies on the velocity of last state, which is neither directly observed by sensor nor involved in optimization. Therefore, the accumulated error of velocity increases with time, degrading the accuracy of state estimation. In addition, LOAM and LeGO-LOAM do not estimate the gravity vector which need to be removed from raw accelerometer measurements. Instead, they obtain the roll, pitch, yaw angle in real time by fusing the magnetometer with the accelerometer and gyroscope measurements in the attitude and heading reference system (AHRS). Then, the obtained roll, pitch and yaw angle are used to remove the gravity vector. However, the magnetometer only exists in AHRS, preventing those systems being used in most hardware platforms with 6-axis IMUs.
The tight-coupled methods \cite{li2021towards, ye2019tightly, qin2020lins, shan2020lio, xu2021fast, xu2022fast, chen2022direct, yuan2022sr} also use IMU measurements to provide motion constraints for ICP so as to improve the accuracy and robustness of state estimation. The LIO joint optimization systems based on the tightly-coupled framework can be mainly categorized into three types: iterated extended Kalman filter (iEKF), \cite{qin2020lins, xu2021fast, xu2022fast}, bundle adjustment (BA) \cite{li2021towards, ye2019tightly, yuan2022sr} and graph optimization \cite{shan2020lio, chen2022direct}. All three types regard pose and velocity as state variables that need to be solved. The prediction of pose and velocity are calculated by integrating IMU measurements to the last state, and the observation of pose is obtained from LiDAR ICP. However, the velocity does not have any observation, therefore, it can only adjust itself according to the result of pose to satisfy the kinematic constraints. In order words, the accuracy of velocity mainly depends on the accuracy of pose. Once the pose is not correct, the velocity would adjust itself to fit the incorrect pose. Although IMU pre-integrations can constrain the velocity, this constrain is not very useful if the velocity of last time is not accurate.

In this paper, we present LIWO, an accurate and robust LiDAR-inertial-wheel (LIW) odometry, which fused constraints from LiDAR, IMU and wheel encoder in a BA based tightly-coupled framework. LIWO integrates the IMU and wheel encoder measurements as the initial pose value of current sweep, and then refine pose by a BA based LIW-optimization module. The wheel encoder can make the initial velocity value of LIW-optimization more reliable, and meanwhile can provide velocity observations to address the limitation of IMU. In addition, compared with \cite{zhang2014loam, shan2018lego} which need AHRS support, our system is compatible with hardware platforms with 6-axis IMUs and thus is much more convenient in practice. Although we utilize an extra wheel encoder sensor, the cost of a wheel encoder sensor is much lower than that of an AHRS. Experimental results on the public dataset $nclt$ \cite{carlevaris2016university} and kaist \cite{jeong2019complex} demonstrate that: 1) our system outperforms existing state-of-the-art LIO systems (i.e., \cite{li2021towards, qin2020lins, shan2020lio, xu2022fast, yuan2022sr} in term of smaller absolute trajectory error (ATE); 2) Compared with a variant of our system which utilizes only IMU pre-integration to provide constraints, our final system which embeds velocity observations from wheel encoder could further improve accuracy and robustness.

To summarize, the main contributions of this work are two folds: 1) We proposed a novel BA based LIW odometry system, which embeds the velocity observations from a wheel encoder into BA based Lidar-inertial optimization framework. Our LIWO outperforms most state-of-the-art LIO systems in terms of accuracy. 2) We have released the source code of this work for the development of the community\footnote{https://github.com/ZikangYuan/liw\_oam}.

The rest of this paper is structured as follows. In Sec. \ref{Related Work}, we briefly discuss the relevant literature. Sec. \ref{Preliminary} provides preliminaries. Sec. \ref{System Overview} illustrates the overview of our system. Sec. \ref{System Details} details each module of our system, followed by experimental evaluation in Sec. \ref{Experiments}. Sec. \ref{Conclusion} concludes the paper.

\section{RELATED WORK}
\label{Related Work}

\textbf{LiDAR-Only Odometry and Mapping.} LiDAR-only odometry and mapping systems rely on geometric information contained in LiDAR points for tracking, and constantly register the new points to the map. LOAM \cite{zhang2014loam, zhang2017low} firstly proposes a complete LiDAR odometry which mainly consists of three steps: 1) Extracting edge and surfaces from raw points; 2) Performing sweep-to-sweep pose estimation at an input sweep frequency; 3) Performing sweep-to-map pose optimization and utilizing the optimized pose to register points to the map at a lower frequency. However, due to huge number of 3D points to be processed, the output frequency of LOAM is low. On the basis of LOAM, LeGO-LOAM \cite{shan2018lego} proposes to cluster raw LiDAR points and then removes clusters with weak geometric structure information to reduce computation. Fast-LOAM \cite{wang2021f} eliminates the sweep-to-sweep pose estimation module and keep only the sweep-to-map pose estimation module to make the system lightweight. CT-ICP \cite{dellenbach2022ct} estimates the state at beginning and ending time of each sweep. By this way, the state at any time during a sweep can be expressed as a function of the beginning state and the ending state. Compared with the previous scheme \cite{zhang2014loam, zhang2017low, shan2018lego, wang2020intensity} which represents the state of a sweep by only the state at beginning time or ending time, CT-ICP is more realistic and meanwhile achieves superior performance.

\textbf{LiDAR-Inertial Odometry.} Almost all LiDAR-inertial odometry systems also include the mapping module, but their mapping is almost exactly the same as LiDAR-only odometry and mapping system. The core change occurred in the odometery module, so we usually omit "mapping" when summarizing them. LiDAR-inertial odometry systems are mainly divided into loosely-coupled framework \cite{zhang2014loam, zhang2017low, shan2018lego} and tightly-coupled framework \cite{li2021towards, ye2019tightly, qin2020lins, shan2020lio, xu2021fast, xu2022fast, chen2022direct, yuan2022sr}. The loose-coupled framework, such as LOAM \cite{zhang2014loam} and LeGO-LOAM \cite{shan2018lego} with an IMU interface, uses IMU measurements to calibrate the motion priors for ICP pose estimation. The tightly-coupled framework uses IMU measurements to provide motion constraints for ICP, so as to improve the accuracy and robustness of pose estimation. According to the type of LiDAR-inertial joint optimization, the tightly-coupled framework can be further divided into iEKF based framework \cite{qin2020lins, xu2021fast, xu2022fast} BA based framework \cite{li2021towards, ye2019tightly, yuan2022sr} and graph optimization based framework \cite{shan2020lio, chen2022direct}. LINs \cite{qin2020lins} firstly fuses 6-axis IMU and 3D LiDAR in an iEKF based framework, where an iEKF is designed to correct the estimated state recursively by generating new feature correspondences in each iteration, and to keep the system computationally tractable. Fast-LIO \cite{xu2021fast} proposes a new method of solving Kalman gain to avoid the calculation of the high-order matrix inversion, and in turn greatly reduce the computational burden. Based on Fast-LIO, Fast-LIO2 \cite{xu2022fast} proposes an ikd-tree algorithm \cite{cai2021ikd}. Compared with the original kd-tree, ikd-tree reduces time cost in building a tree, traversing a tree, removing elements and other operations. LIO-SAM \cite{shan2020lio} formulates LiDAR-inertial odometry as a factor graph. Measurements from LiDAR and IMU are used to provide absolute constraints for each node graph and relative constraints between nodes respectively. DLIO \cite{chen2022direct} builds an internal map by registering dense points to a local submap with a translational and rotational prior generated by a nonlinear motion model. \cite{ye2019tightly} fuses 6-axis IMU and 3D LiDAR in a BA based framework. Besides, to obtain more reliable poses estimation, a rotation-constrained refinement algorithm is proposed to further align the pose with the global map. LiLi-OM \cite{li2021towards} selects the key-sweeps from solid-state LiDAR data, and performs BA based multi-key-sweep joint LiDAR-inertial optimization. However, when the type of LiDAR changes from solid-state to spinning, the time interval between two consecutive key-sweeps becomes longer, and the accumulative error in IMU pre-integration increases. To reduce the accumulative error of IMU pre-integration in BA based framework, our previous work SR-LIO \cite{yuan2022sr} segments and reconstructs raw input sweeps from spinning LiDAR to obtain reconstructed sweeps with higher frequency, and utilizes the reconstructed sweep for state estimation.

\textbf{LiDAR-Inertial-Wheel Odometry.} \cite{zhang2019lidar} is the first approach (without source code released) trying to fuse LiDAR, IMU and wheel in a loosely-coupled extended Kalman filter (EKF) framework. First, the state at a particular time is calculated by LiDAR, IMU and wheel encoder odometer respectively. Then, the three states calculated by the three sensors are integrated into an EKF to obtain the final state. EKF-LOAM \cite{junior2022ekf} which provides the source code also adopts the loosely-coupled EKF framework. Compared with \cite{zhang2019lidar}, \cite{junior2022ekf} uses a simple and lightweight adaptive covariance matrix based on the number of detected geometric features. Different from \cite{zhang2019lidar} and \cite{junior2022ekf}, our LIWO utilizes a tightly-coudpled BA framework, where constraints from LiDAR, IMU and wheel encoder odometer are used together to solve a state. This scheme is more robust, because when the observation of one sensor is unreliable, constraints form other sensors can also ensure the reliability of the final state solution in most cases.

\section{PRELIMINARY}
\label{Preliminary}

\subsection{Coordinate Systems}
\label{Coordinate Systems}

We denote $\left({\cdot}\right)^w$, $\left({\cdot}\right)^l$, $\left({\cdot}\right)^o$ and $\left({\cdot}\right)^k$ as a 3D point in the world coordinates, the LiDAR coordinates, the IMU coordinates and the odometer (i.e., wheel encoder) coordinates respectively. The world coordinate is coinciding with $\left({\cdot}\right)^l$ at the starting position. In all coordinates, the x-axis points forward, the y-axis points to the left, and the z-axis points upward.

We denote the LiDAR coordinates for taking the $i_{th}$ sweep at time $t_i$ as $l_i$ and the corresponding IMU coordinates at $t_i$ as $o_i$, then the transformation matrix (i.e., external parameters) from the LiDAR coordinates $l_i$ to the IMU coordinates $o_i$ is denoted as $\mathbf{T}_{l_i}^{o_i} \in S E(3)$, where $\mathbf{T}_{l_i}^{o_i}$ consists of a rotation matrix $\mathbf{R}_{l_i}^{o_i} \in S O(3)$ and a translation vector $\mathbf{t}_{l_i}^{o_i} \in \mathbb{R}^3$. The external parameters are usually calibrated once offline and remain constant during online pose estimation; therefore, we can represent $\mathbf{T}_{l_i}^{o_i}$ using $\mathbf{T}_{l}^{o}$ for simplicity. Similarity, the transformation from the odometer coordinates to the IMU coordinate is denoted as $\mathbf{T}_{k}^{o}$, which consists of $\mathbf{R}_{k}^{o}$ and $\mathbf{t}_{k}^{o}$.

We use both rotation matrices $\mathbf{R}$ and Hamilton quaternions $\mathbf{q}$ to represent rotation. We primarily use quaternions in state vectors, but rotation matrices are also used for convenience rotation of 3D vectors. $\otimes$ represents the multiplication operation between two quaternions. Finally, we denote $\left(\hat{\cdot}\right)$ as the noisy measurement or estimate of a certain quantity.

In addition to pose, we also estimate the velocity $\mathbf{v}$, the accelerometer bias $\mathbf{b}_{\mathbf{a}}$ and the gyroscope bias $\mathbf{b}_{\boldsymbol{\omega}}$, which are represented uniformly by a state vector:
\begin{equation}
\label{equation2}
	\boldsymbol{x}=\left[\mathbf{t}^T, \mathbf{q}^T, \mathbf{v}^T, \mathbf{b}_{\mathbf{a}}{ }^T, \mathbf{b}_{\boldsymbol{\omega}}{ }^T\right]^T
\end{equation}

\subsection{Sweep State Expression}
\label{Sweep State Expression}

Inspired by CT-ICP \cite{dellenbach2022ct}, we represent the state of a sweep $S$ by: 1) the state at the beginning time $t_b$ of $S$ (e.g., $\boldsymbol{x}_{b}$) and 2) the state at the end time $t_e$ of $S$ (e.g., $\boldsymbol{x}_{e}$). By this way, the state of each point during $\left[t_b, t_e\right]$ can be represented as a function of $\boldsymbol{x}_{b}$ and $\boldsymbol{x}_{e}$. For instance, for a point $\mathbf{p} \in S$ collected at time $t_{\mathbf{p}} \in\left[t_b, t_e\right]$, the state at $t_{\mathbf{p}}$ can be calculated as:
\begin{equation}
\label{equation3}
	\begin{gathered}
		\alpha=\frac{t_{\mathbf{p}}-t_b}{t_e-t_b} \\
		\mathbf{t}_{\mathbf{p}}=(1-\alpha) \mathbf{t}_b+\alpha \mathbf{t}_e \\
		\mathbf{q}_{\mathbf{p}}=\mathbf{q}_b . slerp\left(\alpha, \mathbf{q}_e\right) \\
		\mathbf{v}_{\mathbf{p}}=(1-\alpha) \mathbf{v}_b+\alpha \mathbf{v}_e \\
		\mathbf{b}_{\mathbf{a}_{\mathbf{p}}}=(1-\alpha) \mathbf{b}_{\mathbf{a}_b}+\alpha \mathbf{b}_{\mathbf{a}_e} \\
		\mathbf{b}_{\boldsymbol{\omega}_{\mathbf{p}}}=(1-\alpha) \mathbf{b}_{\boldsymbol{\omega}_b}+\alpha \mathbf{b}_{\boldsymbol{\omega}_e}
	\end{gathered}
\end{equation}
where $slerp\left(\cdot\right)$ is the spherical linear interpolation operator for quaternion.

\subsection{IMU-Odometer Measurement Model}
\label{IMU-Odometer Measurement Model}

The IMU-odometer includes a wheel encoder and an IMU, which consists of an accelerometer and a gyroscope. The raw gyroscope and accelerometer measurements from IMU, i.e., $\hat{\boldsymbol{\omega}}_t$ and $\hat{\mathbf{a}}_t$, are given by:
\begin{equation}
\label{equation4}
	\begin{gathered}
		\hat{\boldsymbol{\omega}}_t=\boldsymbol{\omega}_t+\mathbf{b}_{\boldsymbol{\omega}_t}+\mathbf{n}_{\boldsymbol{\omega}} \\
		\hat{\mathbf{a}}_t=\mathbf{a}_t+\mathbf{b}_{\mathbf{a}_t}+\mathbf{R}_w^t \mathbf{g}^w+\mathbf{n}_{\mathbf{a}}
	\end{gathered}
\end{equation}
IMU measurements, which are measured in the IMU coordinates, combine the force for countering gravity and the platform dynamics, and are affected by acceleration bias $\mathbf{b}_{\mathbf{a}_t}$, gyroscope bias $\mathbf{b}_{\boldsymbol{\omega}_t}$, and additive noise. As mentioned in VINs-Mono \cite{qin2018vins}, the additive noise in acceleration and gyroscope measurements are modeled as Gaussian white noise, $\mathbf{n}_{\mathbf{a}} \sim N\left(\mathbf{0}, \boldsymbol{\sigma}_{\mathbf{a}}^2\right)$, $\mathbf{n}_{\boldsymbol{\omega}} \sim N\left(\mathbf{0}, \boldsymbol{\sigma}_{\boldsymbol{\omega}}^2\right)$. Acceleration bias and gyroscope bias are modeled as random walk, whose derivatives are Gaussian, $\dot{\mathbf{b}}_{\mathbf{a}_t}=\mathbf{n}_{\mathbf{b}_{\mathbf{a}}} \sim N\left(\mathbf{0}, \boldsymbol{\sigma}_{\mathbf{b}_{\mathbf{a}}}^2\right)$, $\dot{\mathbf{b}}_{{\boldsymbol{\omega}}_t}=\mathbf{n}_{\mathbf{b}_{\boldsymbol{\omega}}} \sim N\left(\mathbf{0}, \boldsymbol{\sigma}_{\mathbf{b}_{\boldsymbol{\omega}}}^2\right)$.

The wheel encoder obtains the rotational speed $\tau$ of the shaft according to the pulse received by the counter, and then calculate the speed of the left rear wheel and the right rear wheel according to $\tau$ and the wheel radius $r$:
\begin{equation}
	\label{equation5}
	\begin{gathered}
		\hat{\mathbf{v}}_{left}=\left[\begin{array}{lll}
			\hat{\tau}_{left} r_{left} & 0 & 0
		\end{array}\right]^T \\
		\hat{\mathbf{v}}_{right}=\left[\begin{array}{lll}
			\hat{\tau}_{right} r_{right} & 0 & 0
		\end{array}\right]^T \\
		\hat{\tau}_{left}=\tau_{left}+n_{\tau_{left}}, \hat{\tau}_{right}=\tau_{right}+n_{\tau_{right}}
	\end{gathered}
\end{equation}
where $n_{\tau_{left}}$ and $n_{\tau_{right}}$ are the corresponding zero-mean white Gaussian noises of $\tau_{left}$ and $\tau_{right}$, $\hat{\mathbf{v}}_{left}$ and $\hat{\mathbf{v}}_{right}$ are the measured linear speed of two wheels calculated from $\hat{\tau}_\cdot$ and $r_\cdot$. Then the final measurement model of wheel encoder odometer, which are measured in odometer coordinates, can be defined as:
\begin{equation}
	\label{equation6}
	\begin{gathered}
		\hat{\mathbf{v}}=\frac{\hat{\mathbf{v}}_{left}+\hat{\mathbf{v}}_{right }}{2}+\mathbf{n}_{\mathbf{v}} \\
		\mathbf{n}_{\mathbf{v}}=\left[\begin{array}{ccc}
			\frac{r_{left} n_{\tau_{left}}+r_{right} n_{\tau_{right}}}{2} & 0 & 0
		\end{array}\right]^T
	\end{gathered}
\end{equation}
where $\left[\mathbf{n}_{\mathbf{v}}\right]_x$ is the sum of two zero-mean Gaussian distributions. In theory, $n_{\tau_{left}}$ and $n_{\tau_{right}}$ are correlated. However, we assume that they are independent for simplifying the noise model. Under this assumption, $\left[\mathbf{n}_{\mathbf{v}}\right]_x$ is the sum of two uncorrelated zero-mean Gaussian distributions $\mathbf{n}_{\mathbf{v}} \sim N\left(\mathbf{0}, \boldsymbol{\sigma}_{\mathbf{v}}^2\right)$, which is still a zero-mean Gaussian distribution. This simplified noise model can be processed together with the IMU noise when propagation of covariance.

\section{SYSTEM OVERVIEW}
\label{System Overview}

\begin{figure*}
	\begin{center}
		\includegraphics[scale=0.5]{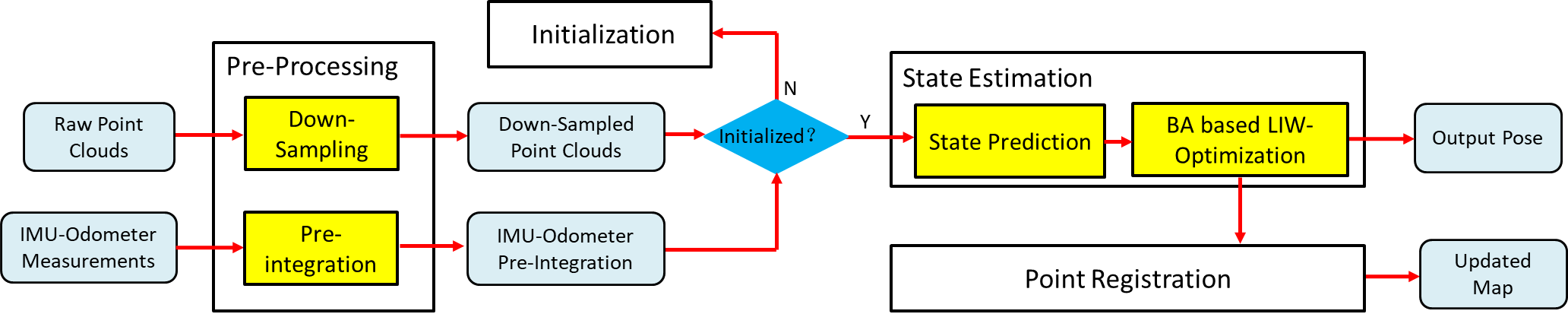}
		\caption{Overview of our LIWO which consists of four main modules: a pre-processing module, an initialization module, a state estimation module and a point registration module.}
		\label{fig1}
	\end{center}
\end{figure*}

Fig. \ref{fig1} illustrates the framework of our LIWO which consists of four main modules: pre-processing, initialization, state estimation and point registration. The pre-processing module down-samples the input raw points, and pre-integrates IMU-odometer measurements at the same frequency of input sweep. The initialization module estimates some state parameters including gravitational acceleration, accelerometer bias, gyroscope bias, and initial velocity. The state estimation module firstly integrates the IMU-odometer measurements to the last state to predict the current state, then performs BA based LIW-optimization to optimize the state of current sweep. Finally, the point registration adds the new points to the map, and deletes the points that are far away.

\section{SYSTEM DETAILS}
\label{System Details}

\subsection{Pre-Processing}
\label{Pre-Processing}

\subsubsection{Down Sampling}
\label{Down Sampling}

Processing a huge number of 3D points yields a high computational cost. To reduce the computational complexity, we down-sample the input points as following. Following CT-ICP \cite{dellenbach2022ct}, we put the points of current input sweep $S_{i+1}$ into a volume with 0.5$\times$0.5$\times$0.5 (unit: m) voxel size, and make each voxel contain only one point, to obtain the down-sampled sweep $P_{i+1}$. This down-sampling strategy ensures that the density distribution of points is uniform in 3D space after down-sampling.

\subsubsection{Pre-Integration}
\label{Pre-Integration}

Typically, the IMU-odometer sends out data at a much higher frequency than the LiDAR. Pre-integration of all IMU-odometer measurements between two consecutive sweeps $S_i$ and $S_{i+1}$ can well summarize the dynamics of the hardware platform from time $t_{e_i}$ to $t_{e_{i+1}}$, where $e_i$ and $e_{i+1}$ are the end time stamp of $S_i$ and $S_{i+1}$ respectively. In this work, we employ the discrete-time quaternion-based derivation of IMU-odometer pre-integration approach \cite{liu2019visual}, and incorporate IMU bias using the method in \cite{qin2018vins}. Specifically, the pre-integrations between $S_i$ and $S_{i+1}$ in the corresponding IMU coordinates $o_{e_i}$ and $o_{e_{i+1}}$, i.e., $\hat{\boldsymbol{\alpha}}_{e_{i+1}}^{e_i}$, $\hat{\boldsymbol{\eta}}_{e_{i+1}}^{e_i}$, $\hat{\boldsymbol{\beta}}_{e_{i+1}}^{e_i}$, and $\hat{\boldsymbol{\gamma}}_{e_{i+1}}^{e_i}$, are calculated, where $\boldsymbol{\alpha}_{e_{i+1}}^{e_i}$, $\boldsymbol{\beta}_{e_{i+1}}^{e_i}$, $\boldsymbol{\gamma}_{e_{i+1}}^{e_i}$ are the pre-integration of translation, velocity, rotation from IMU measurements respectively and $\boldsymbol{\eta}_{e_{i+1}}^{e_i}$ is the pre-integration of translation from gyroscope and wheel encoder odometer measurements. In addition, the Jacobian of pre-integration with respect to bias, i.e., $\mathbf{J}_{\mathbf{b}_{\mathbf{a}}}^{\boldsymbol{\alpha}}$, $\mathbf{J}_{\mathbf{b}_{\boldsymbol{\omega}}}^{\boldsymbol{\alpha}}$, $\mathbf{J}_{\mathbf{b}_{\mathbf{a}}}^{\boldsymbol{\beta}}$, $\mathbf{J}_{\mathbf{b}_{\boldsymbol{\omega}}}^{\boldsymbol{\beta}}$, $\mathbf{J}_{\mathbf{b}_{\boldsymbol{\omega}}}^{\boldsymbol{\gamma}}$, $\mathbf{J}_{\mathbf{b}_{\mathbf{a}}}^{\boldsymbol{\eta}}$ and $\mathbf{J}_{\mathbf{b}_{\boldsymbol{\omega}}}^{\boldsymbol{\eta}}$, are also calculated according to the error state kinematics. 

\subsection{Initialization}
\label{Initialization}

The initialization module aims to estimate all necessary values including initial pose, velocity, gravitational acceleration, accelerometer bias and gyroscope bias, for subsequent state estimation. Similar as our previous work SR-LIO \cite{yuan2022sr}, we adopt motion initialization and static initialization for handheld devices and vehicle-mounted devices respectively. Please refer to \cite{yuan2022sr} for more details about our initialization module.

\subsection{State Estimation}
\label{State Estimation}

\subsubsection{State Prediction}
\label{State Prediction}

When every new down-sampled sweep $P_{i+1}$ completes, we use IMU-odometer measurements to predict the state at the beginning time stamp of $P_{i+1}$ (i.e., $\boldsymbol{x}_{b_{i+1}}^w$) and the state at the end time stamp of $P_{i+1}$ (i.e., $\boldsymbol{x}_{e_{i+1}}^w$) to provide the prior motion for LIW-optimization. Specifically, the predicted state $\boldsymbol{x}_{b_{i+1}}^w$ (i.e., $\mathbf{t}_{b_{i+1}}^w$, $\mathbf{R}_{b_{i+1}}^w$, $\mathbf{v}_{b_{i+1}}^w$, $\mathbf{b}_{\mathbf{a}_{b_{i+1}}}$ and $\mathbf{b}_{\boldsymbol{\omega}_{b_{i+1}}}$) is assigned as:
\begin{equation}
\label{equation7}
	\boldsymbol{x}_{b_{i+1}}^w=\boldsymbol{x}_{e_i}^w
\end{equation}
and $\boldsymbol{x}_{e_{i+1}}^w$ (i.e., $\mathbf{t}_{e_{i+1}}^w$, $\mathbf{R}_{e_{i+1}}^w$, $\mathbf{v}_{e_{i+1}}^w$, $\mathbf{b}_{\mathbf{a}_{e_{i+1}}}$ and $\mathbf{b}_{\boldsymbol{\omega}_{e_{i+1}}}$) is calculated as:
\begin{equation}
\label{equation8}
	\begin{gathered}
		\mathbf{R}_{n+1}^w=\mathbf{R}_n^w Exp\left(\left(\frac{\hat{\boldsymbol{\omega}}_n+\hat{\boldsymbol{\omega}}_{n+1}}{2}-\mathbf{b}_{\boldsymbol{\omega}_{e_i}}\right) \delta t\right) \\
		\mathbf{v}_{n+1}^w=\mathbf{R}_{n+1}^w \mathbf{R}_k^l \hat{\mathbf{v}}_{n+1} \\
		\mathbf{t}_{n+1}^w=\mathbf{t}_n^w+\mathbf{v}_n^w \delta t+\frac{1}{2}\left(\frac{\hat{\mathbf{a}}_n+\hat{\mathbf{a}}_{n+1}}{2}-\mathbf{b}_{\mathbf{a}_{e_i}}-\mathbf{R}_n^w \mathbf{g}^w\right) \delta t^2
	\end{gathered}
\end{equation}
where $\hat{\boldsymbol{\omega}}_{\cdot}$, $\hat{\mathbf{a}}_{\cdot}$ and $\hat{\mathbf{v}}_{\cdot}$ are the measurements from IMU gyroscope, IMU accelerometer and wheel encoder, $\mathbf{g}^w$ is the gravitational acceleration in the world coordinates, $n$ and $n+1$ are two time instants of obtaining an IMU-odometer measurements during $\left[t_{e_i}, t_{e_{i+1}}\right]$, $\delta t$ is the time interval between $n$ and $n+1$. We iteratively increase $n$ from $0$ to $\left(t_{e_{i+1}}-t_{e_i}\right) / \delta t$ to obtain $\boldsymbol{x}_{e_{i+1}}^w$. When $n=0$, $\boldsymbol{x}_{n}^w=\boldsymbol{x}_{e_i}^w$. For $\mathbf{b}_{\mathbf{a}_{e_{i+1}}}$ and $\mathbf{b}_{\boldsymbol{\omega}_{e_{i+1}}}$, we set the predicted values of them by: $\mathbf{b}_{\mathbf{a}_{e_{i+1}}}=\mathbf{b}_{\mathbf{a}_{e_i}}$ and $\mathbf{b}_{\boldsymbol{\omega}_{e_{i+1}}}=\mathbf{b}_{\boldsymbol{\omega}_{e_i}}$.

\subsubsection{BA based LIW-Optimization}
\label{BA based LIW-Optimization}

We jointly utilize measurements of the LiDAR, inertial and wheel encoder to optimize the beginning state (i.e., $\boldsymbol{x}_{b_{i+1}}^w$) and the end state (i.e., $\boldsymbol{x}_{e_{i+1}}^w$) of the current sweep $P_{i+1}$, where the variable vector is expressed as:
\begin{equation}
\label{equation9}
	\boldsymbol{\chi}=\left\{\boldsymbol{x}_{b_{i+1}}^w, \boldsymbol{x}_{e_{i+1}}^w\right\}
\end{equation}
\textbf{Residual from the LiDAR constraint.} As proposed in Fast-LIO2 \cite{xu2022fast}, directly building point-to-plane residuals is more accurate and robust than using both edge and surface features. Therefore, we followed their scheme in this work. For a point $\mathbf{p}$, we first project $\mathbf{p}$ to the world coordinates to obtain $\mathbf{p}^w$, and then find 20 nearest points around $\mathbf{p}^w$ from the volume. To search for the nearest neighbor of $\mathbf{p}^w$, we only search in the voxel $V$ to which $\mathbf{p}^w$ belongs, and the 8 voxels adjacent to $V$. The 20 nearest points are used to fit a plane with a normal $\mathbf{n}$ and a distance $d$. Accordingly, we can build the point-to-plane residual $r^{\mathbf{p}}$ for $\mathbf{p}$ as:
\begin{equation}
\label{equation10}
	\begin{gathered}
		r^{\mathbf{p}}=\omega_{\mathbf{p}}\left(\mathbf{n}^T \mathbf{p}^w+d\right) \\
		\mathbf{p}^w=\mathbf{q}_{\mathbf{p}}^w \mathbf{p}+\mathbf{t}_{\mathbf{p}}^w \\
		\alpha=\frac{t_{\mathbf{p}}-t_{b_{i+1}}}{t_{e_{i+1}}-t_{b_{i+1}}} \\
		\mathbf{t}_{\mathbf{p}}^w=(1-\alpha) \mathbf{t}_{b_{i+1}}^w+\alpha \mathbf{t}_{e_{i+1}}^w \\
		\mathbf{q}_{\mathbf{p}}^w=\mathbf{q}_{b_{i+1}}^w . slerp\left(\alpha, \mathbf{q}_{e_{i+1}}^w\right)
	\end{gathered}
\end{equation}
where $\omega_{\mathbf{p}}$ is a weight parameter defined by \cite{dellenbach2022ct}, $\mathbf{q}_{b_{i+1}}^w$ and $\mathbf{q}_{e_{i+1}}^w$ are the rotation with respect to $\left({\cdot}\right)^w$ at $t_{b_{i+1}}$ and $t_{e_{i+1}}$ respectively, $\mathbf{t}_{b_{i+1}}^w$ and $\mathbf{t}_{e_{i+1}}^w$ are the translation with respect to $\left({\cdot}\right)^w$ at $t_{b_{i+1}}$ and $t_{e_{i+1}}$ respectively. Both $\mathbf{q}_{b_{i+1}}^w$, $\mathbf{q}_{e_{i+1}}^w$, $\mathbf{t}_{b_{i+1}}^w$, $\mathbf{t}_{e_{i+1}}^w$ are variables to be refined, and the initial value of them are obtained from Sec. \ref{State Prediction}.

\textbf{Residual from the IMU-odometer constraint.} Considering the IMU-odometer measurements during $\left[t_{b_{i+1}}, t_{e_{i+1}}\right]$, according to pre-integration introduced in \cite{liu2019visual}, the residual for pre-integrated IMU-odometer measurements can be computed as:
\begin{equation}
\label{equation11}
	\begin{gathered}
		{{\mathbf{r}_o}_{{e_{i+1}}}^{b_{i+1}}}= \\
		{\left[\begin{array}{c}
				\mathbf{R}_w^{b_{i+1}}\left(\mathbf{t}_{e_{i+1}}^w-\mathbf{t}_{b_{i+1}}^w+\frac{1}{2} \mathbf{g}^w \Delta t^2-\mathbf{v}_{b_{i+1}}^w \Delta t\right)-\hat{\boldsymbol{\alpha}}_{e_{i+1}}^{e_i} \\
				\mathbf{R}_w^{b_{i+1}}\left(\mathbf{v}_{e_{i+1}}^w+\mathbf{g}^w \Delta t-\mathbf{v}_{b_{i+1}}^w\right)-\hat{\boldsymbol{\beta}}_{e_{i+1}}^{e_i} \\
				2\left[\mathbf{q}_{b_{i+1}}^{{w}^{-1}} \otimes \mathbf{q}_{e_{i+1}}^w \otimes\left(\hat{\boldsymbol{\gamma}}_{e_{i+1}}^{e_i}\right)^{-1}\right]_{x y z} \\
				\mathbf{R}_w^{b_{i+1}}\left(\mathbf{t}_{e_{i+1}}^w-\mathbf{t}_{b_{i+1}}^w\right)-\mathbf{t}_k^o+\mathbf{R}_w^{b_{i+1}} \mathbf{R}_{e_{i+1}}^w \mathbf{t}_k^o-\hat{\boldsymbol{\eta}}_{e_{i+1}}^{e_i} \\
				\mathbf{b}_{\mathbf{a}_{i+1}}-\mathbf{b}_{\mathbf{a}_i} \\
				\mathbf{b}_{\boldsymbol{\omega}_{i+1}}-\mathbf{b}_{\boldsymbol{\omega}_i}
			\end{array}\right]}
	\end{gathered}
\end{equation}
where $[\cdot]_{x y z}$ extracts the vector part of a quaternion $\mathbf{q}$ for error state representation. At the end of each iteration, we update $\left[\hat{\boldsymbol{\alpha}}_{e_{i+1}}^{e_i}, \hat{\boldsymbol{\beta}}_{e_{i+1}}^{e_i}, \hat{\boldsymbol{\gamma}}_{e_{i+1}}^{e_i}, \hat{\boldsymbol{\eta}}_{e_{i+1}}^{e_i}\right]^T$ with the first order Jacobian approximation \cite{qin2018vins}.

\textbf{Residual from the velocity observation constraint.} As mentioned in Sec. \ref{Introduction}, the existing LIO systems lack of velocity observations to constrain the velocity during optimization. In our system, we utilized the measurements from wheel encoder as the observation to constrain the velocity:
\begin{equation}
\label{equation12}
	\mathbf{r}_w=\left[\begin{array}{l}
		\mathbf{r}_{w_{b_{i+1}}} \\
		\mathbf{r}_{w_{e_{i+1}}}
	\end{array}\right]=\left[\begin{array}{l}
		\mathbf{v}_{b_{i+1}}^w-\mathbf{R}_{b_{i+1}}^w \mathbf{R}_k^l \hat{\mathbf{v}}_{b_{i+1}} \\
		\mathbf{v}_{e_{i+1}}^w-\mathbf{R}_{e_{i+1}}^w \mathbf{R}_k^l \hat{\mathbf{v}}_{e_{i+1}}
	\end{array}\right]
\end{equation}
where $\mathbf{R}_k^l$ is the rotation from the wheel encoder to LiDAR, $\hat{\mathbf{v}}_{b_{i+1}}$ and $\hat{\mathbf{v}}_{e_{i+1}}$ are the velocity measurements from the wheel encoder at $t_{b_{i+1}}$ and $t_{e_{i+1}}$ respectively.

\textbf{Residual from the consistency constraint.} According to CT-ICP \cite{dellenbach2022ct}, $\boldsymbol{x}_{b_{i+1}}^w$ and $\boldsymbol{x}_{e_{i}}^w$ are two states at the same time stamp $t_{b_{i+1}}$($t_{e_i}$). Logically, $\boldsymbol{x}_{e_{i}}^w$ and $\boldsymbol{x}_{b_{i+1}}^w$ should be the same. Therefore, we build the consistency residual as follow:
\begin{equation}
\label{equation13}
	\mathbf{r}_c=\left[\begin{array}{c}
		\mathbf{r}_c^{\mathbf{t}} \\
		\mathbf{r}_c^{\mathbf{q}} \\
		\mathbf{r}_c^{\mathbf{v}} \\
		\mathbf{r}_c^{\mathbf{b}_{\mathbf{a}}} \\
		\mathbf{r}_c^{\mathbf{b_{\boldsymbol{\omega}}}}
	\end{array}\right]=\left[\begin{array}{c}
		\mathbf{t}_{b_{i+1}}^w-\mathbf{t}_{e_i}^w \\
		2\left[{\mathbf{q}_{e_i}^{w}}^{-1} \otimes \mathbf{q}_{b_{i+1}}^w\right]_{x y z} \\
		\mathbf{v}_{b_{i+1}}^w-\mathbf{v}_{e_i}^w \\
		\mathbf{b}_{\mathbf{a}_{b_{i+1}}}-\mathbf{b}_{\mathbf{a}_{e_i}} \\
		\mathbf{b}_{\boldsymbol{\omega}_{b_{i+1}}}-\mathbf{b}_{\boldsymbol{\omega}_{e_i}}
	\end{array}\right]
\end{equation}
where $\mathbf{t}_{b_{i+1}}^w$, $\mathbf{q}_{b_{i+1}}^w$, $\mathbf{v}_{b_{i+1}}^w$, $\mathbf{b}_{\mathbf{a}_{b_{i+1}}}$, $\mathbf{b}_{\boldsymbol{\omega}_{b_{i+1}}}$ are varibales to be optimized.

By minimizing the sum of point-to-plane residuals, the IMU-odometer pre-integration residuals, the velocity observation residuals and the consistency residuals, we obtain maximum a posteriori estimation as:
\begin{equation}
\label{equation14}
\begin{gathered}
	\boldsymbol{\chi}=\min _{\boldsymbol{\chi}} \\ \left\{\rho\left(\sum_{\mathbf{p} \in P_{i+1}}\left\|r^{\mathbf{p}}\right\|_{\mathbf{P}_L}^2+\left\|{\mathbf{r}_o}_{e_{i+1}}^{b_{i+1}}\right\|_{\mathbf{P}_{e_{i+1}}^{e_i}}^2+\left\|\mathbf{r}_w\right\|^2+\left\|\mathbf{r}_c\right\|^2\right)\right\}
\end{gathered}
\end{equation}
where $\rho$ is the Huber kernel to eliminate the influence of outlier residuals. $\mathbf{P}_{e_{i+1}}^{e_i}$ is the covariance matrix of pre-integrated IMU-odometer measurements. The inverse of $\mathbf{P}_{e_{i+1}}^{e_i}$ is utilized as the weight of IMU pre-integration residuals. $\mathbf{P}_L$ is a constant (e.g., 0.001 in our system) to indicate the reliability of the point-to-plane residuals. The inverse of $\mathbf{P}_L$ is utilized as the weight of point-to-plane residuals. After finishing LIW-optimization, we selectively add the points of current sweep to the map.

\subsection{Point Registration}
\label{Point Registration}

Similar as CT-ICP \cite{dellenbach2022ct}, the cloud map is stored in a volume, and the size of each voxel is 1.0$\times$1.0$\times$1.0 (unit: m). Each voxel contains a maximum of 20 points. When the state of the current down-sampled sweep $P_{i+1}$ has been estimated, we transform $P_{i+1}$ to the world coordinate system $\left({\cdot}\right)^w$, and add the transformed points into the volume map. If a voxel already has 20 points, the new points cannot be added to it.

\section{EXPERIMENTS}
\label{Experiments}

\begin{table}[]
\begin{center}
\caption{Datasets for Evaluation}
\label{table1}
	\begin{tabular}{c|cc|cc|cc}
		\hline
		\multirow{2}{*}{} & \multicolumn{2}{c|}{LiDAR} & \multicolumn{2}{c|}{IMU} & \multicolumn{2}{c}{Wheel encoder} \\ \cline{2-7} 
		& Line        & Rate         & Type        & Rate       & Type            & Rate            \\ \hline
		$nclt$              & 32          & 10\,Hz        & 9-axis      & 100\,Hz     & speed           & 10\,Hz           \\
		$kaist$             & 16          & 10\,Hz        & 9-axis      & 200\,Hz     & pulse           & 100\,Hz          \\ \hline
	\end{tabular}
\end{center}
\end{table}

We evaluate our LIWO on the public datasets $nclt$ \cite{carlevaris2016university} and $kaist$ \cite{jeong2019complex}. $nclt$ is a large-scale, long-term autonomous unmanned ground vehicle dataset collected in the University of Michigans North Campus. The $nclt$ dataset contains a full data stream from a Velodyne HDL-32E LiDAR, 50\,Hz data from Microstrain MS25 IMU and 10\,Hz data from Segway vehicle platform’s wheel encoder. The $nclt$ dataset has a much longer duration and amount of data than other datasets and contains several open scenes, such as a large open parking lot. In addition, 50\,Hz IMU measurements cannot meet the requirements of some systems (e.g., LIO-SAM \cite{shan2020lio}). Therefore, we increase the frequency of the IMU to 100\,Hz by interpolation.

\begin{table}[]
\begin{center}
	\caption{Datasets of All Sequences for Evaluation}
	\label{table2}
	\begin{tabular}{cccc}
		\hline
		& Name       & \begin{tabular}[c]{@{}c@{}}Duration\\ (min:sec)\end{tabular} & \begin{tabular}[c]{@{}c@{}}Distamce\\ (km)\end{tabular} \\ \hline
		\textit{nclt\_1}  & 2012-01-08 & 92:16                                                        & 6.4                                                     \\
		\textit{nclt\_2}  & 2012-01-15 & 110:46                                                       & 7.5                                                     \\
		\textit{nclt\_3}  & 2012-01-22 & 86:11                                                        & 6.1                                                     \\
		\textit{nclt\_4}  & 2012-02-02 & 96:39                                                        & 6.2                                                     \\
		\textit{nclt\_5}  & 2012-02-18 & 88:19                                                        & 6.2                                                     \\
		\textit{nclt\_6}  & 2012-03-17 & 81:51                                                        & 5.8                                                     \\
		\textit{nclt\_7}  & 2012-05-11 & 83:36                                                        & 6.0                                                     \\
		\textit{nclt\_8}  & 2012-05-26 & 97:23                                                        & 6.3                                                     \\
		\textit{nclt\_9}  & 2012-06-15 & 55:10                                                        & 4.1                                                     \\
		\textit{nclt\_10} & 2012-08-04 & 79:27                                                        & 5.5                                                     \\
		\textit{nclt\_11} & 2012-08-20 & 88:44                                                        & 6.0                                                     \\
		\textit{nclt\_12} & 2013-09-28 & 76:40                                                        & 5.6                                                     \\
		\textit{kaist\_1} & urban\_08  & 5:07                                                         & 1.56                                                    \\
		\textit{kaist\_2} & urban\_13  & 24:14                                                        & 2.36                                                    \\
		\textit{kaist\_3} & urban\_14  & 29:06                                                        & 8.20                                                    \\ \hline
	\end{tabular}
\end{center}
\end{table}

The $kaist$ dataset is collected with a human-driving robocar on a variety of longer and larger environments. The robocar has two 10\,Hz Velodyne VLP-16, 200\,Hz Ssens MTi-300 IMU and 100\,Hz RLS LM13 wheel encoder. Two 3D LiDARs are tilted by approximately $45^{\circ}$. For point clouds, we utilize the data from both two 3D LiDARs. The datasets’ information, including the sensors’ type and data rate, are illustrated in Table \ref{table1}. As both datasets utilize the vehicle platform, we employ static initialization in our system. Details of all the 15 sequences used in this section, including name, duration, and distance, are listed in Table II. For both datasets, we utilize the universal evaluation metric – absolute translational error (ATE) for pose accuracy evaluation. A consumer-level computer equipped with an Intel Core i7-12700 and 32 GB RAM is used for all experiments.

\subsection{Comparison with the State-of-the-Arts}
\label{Comparison with the State-of-the-Arts}

\begin{table}[]
\begin{center}
	\caption{RMSE of ATE Comparison of State-of-the-art (Unit: m)}
	\label{table3}
	\begin{tabular}{c|ccccc|c}
		\hline
		& \begin{tabular}[c]{@{}c@{}}LiLi-\\ OM\end{tabular} & \begin{tabular}[c]{@{}c@{}}LIO-\\ SAM\end{tabular} & LINs & \begin{tabular}[c]{@{}c@{}}Fast-\\ LIO2\end{tabular} & \begin{tabular}[c]{@{}c@{}}SR-\\ LIO\end{tabular} & Ours           \\ \hline
		nclt\_1  & 60.98                                              & 1.71                                               & $\times$    & \textbf{1.34}                                        & 1.55                                              & 1.42           \\
		nclt\_2  & 127.5                                              & 2.12                                               & $\times$    & 1.65                                                 & 1.53                                              & \textbf{1.46}  \\
		nclt\_3  & 42.32                                              & 9.70                                               & $\times$    & 1.91                                                 & 6.72                                              & \textbf{1.20}  \\
		nclt\_4  & 40.14                                              & \textbf{1.45}                                               & $\times$    & 1.95                                                 & 1.57                                              & \textbf{1.45}  \\
		nclt\_5  & $\times$                                                  & 5.66                                               & $\times$    & 4.37                                                 & 1.46                                              & \textbf{1.44}  \\
		nclt\_6  & 146.2                                              & $\times$                                                  & $\times$    & 6.11                                                 & 2.07                                              & \textbf{1.52}  \\
		nclt\_7  & 89.98                                              & $\times$                                                  & $\times$    & 2.42                                                 & 1.87                                              & \textbf{1.79}  \\
		nclt\_8  & 43.46                                              & $\times$                                                  & $\times$    & 2.62                                                 & 2.04                                              & \textbf{1.41}  \\
		nclt\_9  & 82.66                                              & 1.51                                               & $\times$    & 2.09                                                 & 2.00                                              & \textbf{1.31}  \\
		nclt\_10 & 96.87                                              & 2.26                                               & $\times$    & 2.43                                                 & 2.15                                              & \textbf{1.46}  \\
		nclt\_11 & 207.1                                              & 10.81                                              & $\times$    & 2.29                                                 & 1.97                                              & \textbf{1.33}  \\
		nclt\_12 & 1137.8                                             & $\times$                                                  & $\times$    & 2.91                                                 & 2.32                                              & \textbf{1.55}  \\
		kaist\_1 & $\times$                                                  & $\times$                                                  & $\times$    & 16.27                                                & 3.17                                              & \textbf{2.87}  \\
		kaist\_2 & $\times$                                                  & $\times$                                                  & $\times$    & $\times$                                                    & $\times$                                                 & \textbf{2.68}  \\
		kaist\_3 & $\times$                                                  & $\times$                                                  & $\times$    & $\times$                                                    & $\times$                                                 & \textbf{55.95} \\ \hline
	\end{tabular}
\end{center}
\end{table}

\begin{table}[]
	\begin{center}
		\caption{Ablation Study of Embedding Sensors on RMSE of ATE (Unit: m)}
		\label{table4}
		\begin{tabular}{c|cc|c}
			\hline
			& LiDAR-only & LIO & LIWO \\ \hline
			nclt\_1  & $\times$                                                     & $\times$                                                 & \textbf{1.42}                                      \\
			nclt\_2  & $\times$                                                     & \textbf{1.43}                                     & 1.46                                               \\
			nclt\_3  & $\times$                                                     & $\times$                                                 & \textbf{1.20}                                      \\
			nclt\_4  & $\times$                                                     & \textbf{1.36}                                     & 1.45                                               \\
			nclt\_5  & $\times$                                                     & $\times$                                                 & \textbf{1.44}                                      \\
			nclt\_6  & $\times$                                                     & 1.73                                              & \textbf{1.52}                                      \\
			nclt\_7  & $\times$                                                     & $\times$                                                 & \textbf{1.79}                                      \\
			nclt\_8  & $\times$                                                     & 1.65                                              & \textbf{1.41}                                      \\
			nclt\_9  & $\times$                                                     & 1.72                                              & \textbf{1.31}                                      \\
			nclt\_10 & $\times$                                                     & $\times$                                                 & \textbf{1.46}                                      \\
			nclt\_11 & $\times$                                                     & 76.69                                             & \textbf{1.33}                                      \\
			nclt\_12 & $\times$                                                     & 32.84                                             & \textbf{1.55}                                      \\
			kaist\_1 & $\times$                                                     & $\times$                                     & \textbf{2.87}                                               \\
			kaist\_2 & $\times$                                                     & $\times$                                                 & \textbf{2.68}                                      \\
			kaist\_3 & $\times$                                                     & $\times$                                                 & \textbf{55.95}                                     \\ \hline
		\end{tabular}
	\end{center}
\end{table}

We compare our LIWO with five state-of-the-art LIO systems: i.e., LiLi-OM \cite{li2021towards}, LIO-SAM \cite{shan2020lio}, LINs \cite{qin2020lins}, Fast-LIO2 \cite{xu2022fast} and SR-LIO \cite{yuan2022sr}. It is necessary to emphasize that the LiDAR of $nclt$ takes 130$\sim$140ms to complete a 360deg sweep (i.e., the frequency of a sweep is about 7.5\,Hz), and SR-LIO needs 360deg sweeps as input. Therefore, for SR-LIO, we package sweeps at 7.5\,Hz from the full LiDAR data stream of $nclt$. For fair comparison, we obtain the results of the above systems based on the source code provided by the authors. In addition to the aforementioned LIO systems, there are also a few LIW odometry systems (e.g., \cite{zhang2019lidar} and EKF-LOAM \cite{liu2019visual}). However, \cite{zhang2019lidar} did not provide the source code, and we fail to configure the environment of \cite{liu2019visual} based on their guidance on the github. In addition, the paper of \cite{zhang2019lidar} and \cite{liu2019visual} did not report their ATE results on $nclt$ and $kaist$, but only test on their own datasets. Therefore, the results of \cite{zhang2019lidar} and \cite{liu2019visual} are not included in Table \ref{table3}.

\begin{table}[]
	\begin{center}
		\caption{Time Consumption Per Sweep (Unit: ms)}
		\label{table5}
		\begin{tabular}{c|cc|c}
			\hline
			& LIW-Optimization & Point Registration & Total \\ \hline
			nclt\_1  & 51.1             & 10.3               & 62.7  \\
			nclt\_2  & 51.2             & 13.9               & 66.4  \\
			nclt\_3  & 54.7             & 9.2                & 65.2  \\
			nclt\_4  & 51.8             & 9.7                & 63.0  \\
			nclt\_5  & 53.2             & 10.1               & 64.8  \\
			nclt\_6  & 51.6             & 9.6                & 62.6  \\
			nclt\_7  & 52.3             & 9.7                & 63.3  \\
			nclt\_8  & 54.5             & 9.5                & 65.3  \\
			nclt\_9  & 55.7             & 8.0                & 65.1  \\
			nclt\_10 & 54.5             & 9.2                & 65.1  \\
			nclt\_11 & 54.8             & 9.6                & 65.7  \\
			nclt\_12 & 54.1             & 9.8                & 65.2  \\
			kaist\_1 & 71.9             & 4.2                & 77.2  \\
			kaist\_2 & 76.5             & 3.7                & 81.0  \\
			kaist\_3 & 68.0             & 4.2                & 73.2  \\ \hline
		\end{tabular}
	\end{center}
\end{table}

\begin{figure*}
	\begin{center}
		\includegraphics[scale=0.7]{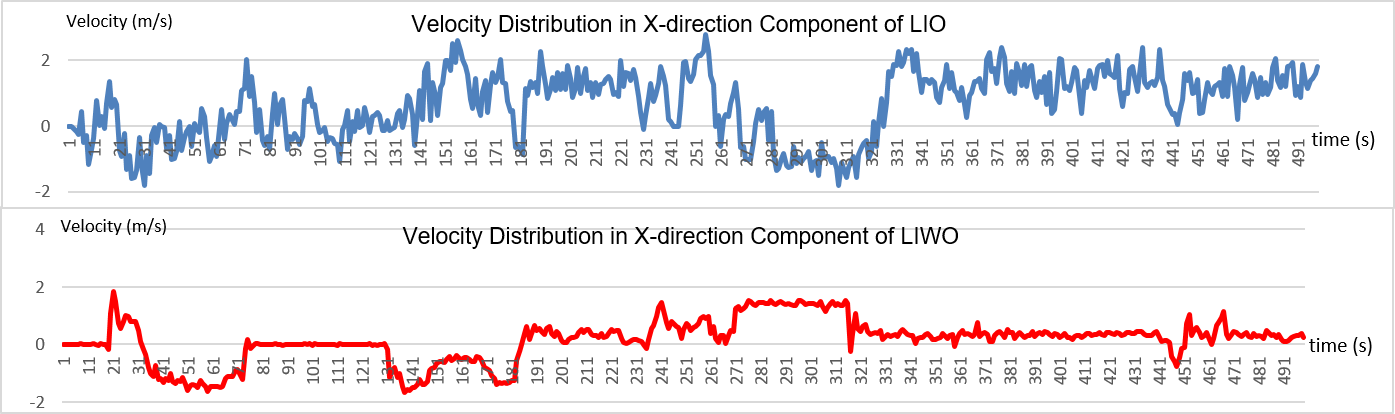}
		\caption{The velocity distribution in the X-direction component of LIO and LIWO on the sequence $nclt\_9$. Compared to the high frequency oscillation curve of LIO, the curve of LIWO is much smoother. This shows that the accuracy of velocity estimation is greatly improved after embedding the velocity observation, because the velocity of a moving vehicle should be continuous and smooth in theory, but not oscillating at high frequency.}
		\label{fig2}
	\end{center}
\end{figure*}

\begin{figure}
	\begin{center}
		\includegraphics[scale=0.33]{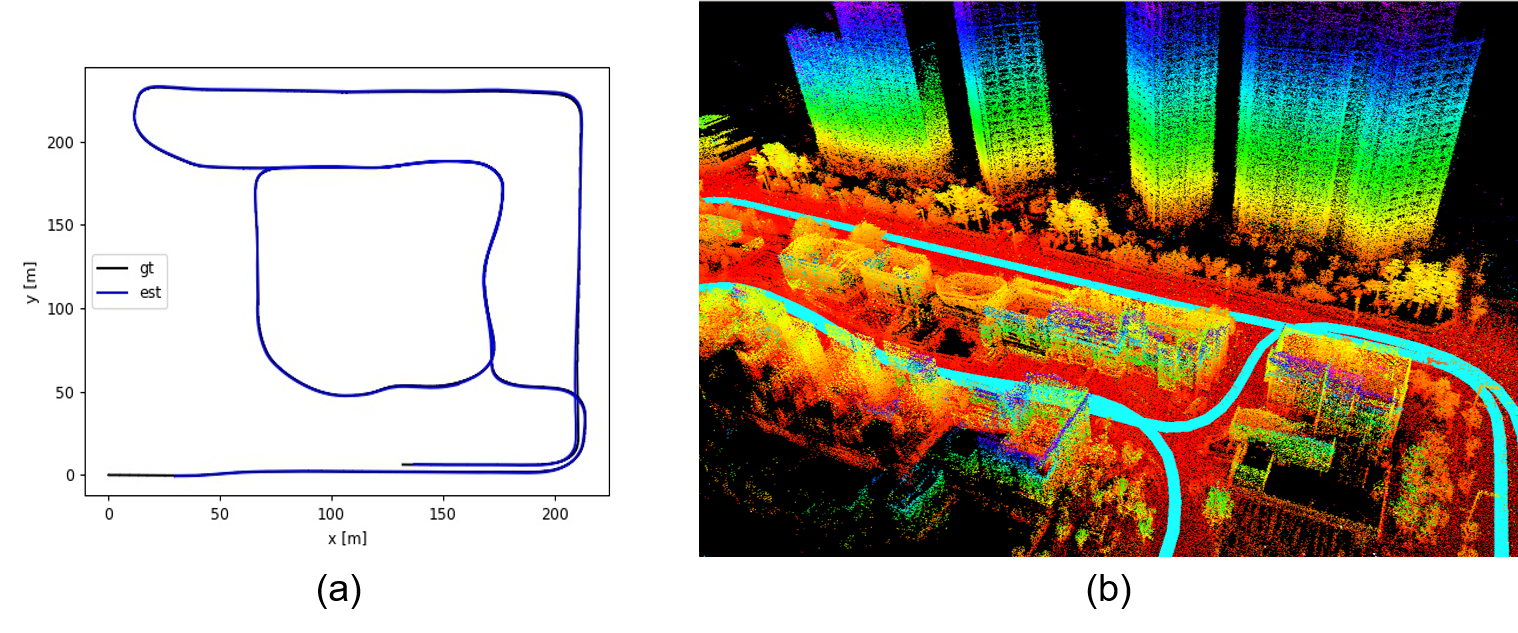}
		\caption{(a) is the comparison result between our estimated trajectory and ground truth on the exemplar sequence $kaist\_1$. (b) is the local point cloud map of $kaist\_1$.}
		\label{fig3}
	\end{center}
\end{figure}

Results in Table \ref{table3} demonstrate that our LIWO outperforms the state-of-the-art LIO systems for almost all sequences in terms of smaller ATE. “$\times$” means the system fails to run the entire sequence. Except for our system, other systems break down on several sequences (especially the $kaist$ sequences), which demonstrate that our method is robust under challenging scenes. It is necessary to emphasize that the Segway vehicle platform enters a long indoor corridor through a door from the outdoor scene at the end of some sequences of $nclt$, yielding significant scene changes. This large differences in scenes produce great difficulties for ICP point cloud registration, and hence almost all LIO systems break down here. Therefore, we omit the test for these cases which usually locate at the end of the sequences.

\subsection{Ablation Study of Embedding Sensors}
\label{Ablation Study of Embedding Sensors}

In this section, we examine the impact of embedding IMU into our LiDAR-only system and embedding the wheel encoder into our BA based LIO system. To this end, we evaluate ATE of the estimated pose under the following three configurations: 1) using only the LiDAR point-to-plane residuals and consistency residuals to provide constraints for state estimation. 2) concurrently using LiDAR point-to-plane residuals, IMU pre-integration residuals and consistency residuals to provide constraints for state estimation. 3) concurrently using LiDAR point-to-plane residuals, IMU-odometer pre-integration residuals, velocity observation residuals and consistency residuals to provide constraints for state estimation. Table \ref{table4} shows the comparison results. Although the accuracy of LIWO is not the best on $nclt\_1$ and $nclt\_4$, we are very close to the best accuracy. On other sequences, embedding wheel encoder can achieve the best performance. In addition, both LiDAR-only and LIO break down on many sequences, while LIWO can run successfully on all sequences. This demonstrates that embedding a wheel encoder can greatly improve the robustness of BA based LIO framework. 

\subsection{Time Consumption}
\label{Time Consumption}

We evaluate the runtime breakdown (unit: ms) of our system for all sequences. In general, the most time-consuming modules are the BA based LIW-optimization module, and the point registration module. Therefore, for each sequence, we test the time cost of above two modules, and the total time for handling a sweep.

Results in Table \ref{table5} show that our LIWO takes 60$\sim$80ms to handle a sweep, while the time interval of two consecutive input sweeps is 100ms. That means our system can not only run in real time, but also save 20$\sim$40ms per sweep.

\subsection{Evaluation of Velocity}
\label{Evaluation of Velocity}

Introducing the wheel encoder measurements, which provides the velocity observation, could yield the greatest accuracy improvement for the estimated velocity, as the raw LIO framework does not have any velocity observations. In practice, it is difficult for us to obtain the ground truth value of velocity. However, we can still make a stereotypical evaluation of the accuracy of velocity based on the kinematic attempt. In theory, the velocity of a moving vehicle should be continuous and smooth, but not oscillating at a high frequency. Therefore, the smoother the curve of a velocity function with respect to time, the better the curve fits the kinematics attempt. As illustrated in Fig. \ref{fig2}, compared to the high frequency oscillation curve of LIO, the curve of LIWO is much smoother. This shows that the accuracy of velocity estimation is greatly improved after embedding velocity observation.

\subsection{Visualization for map}
\label{Visualization for map}

We also visualize the trajectories and point cloud maps estimated by our LIWO. The comparison result between our estimated trajectory and ground truth of the sequence $kaist\_1$ is shown in Fig. \ref{fig3} (a), where our estimated trajectories and ground truth almost exactly coincide. Fig. \ref{fig3} (b) shows sufficient accuracy for some local structures, where the distribution of the points is also uniform.

\section{CONCLUSION}
\label{Conclusion}

In this work, we proposed LIWO, which is an accurate and robust BA based LiDAR-inertial-wheel framework for state estimation in real time. Compared with existing LIO systems, the involvement of a wheel encoder provides velocity measurements as an extra observation, which can greatly enhance the accuracy and robustness of state estimation. Experiment results on the $nclt$ and $kaist$ datasets demonstrate the high accuracy of our LIWO.

It is noteworthy that so far LIWO is only a basic open-sourced LiDAR-inertial-wheel framework. Several practical problems (e.g., the speed inconsistency of the left and right wheels during turns, the observation failure of wheel slippage) have not been considered in the current framework. These problems will be improved in our future work.

\bibliographystyle{IEEEtrans}
\bibliography{IEEEabrv,IEEEExample}

\end{document}